\documentclass[conference]{IEEEtran}
\IEEEoverridecommandlockouts
\usepackage{cite}
\usepackage{amsmath,amssymb,amsfonts}
\usepackage{makecell}
\usepackage{algorithmic}
\usepackage{textcomp}
\usepackage[acronym]{glossaries}
\usepackage[numbers,sort]{natbib}
\usepackage{caption}
\usepackage[dvipsnames]{xcolor}
\usepackage[most]{tcolorbox}
\usepackage{lipsum}
\usepackage{subfigure}
\usepackage{fancyvrb}
\usepackage{siunitx}

\makeglossaries

\newglossaryentry{commit}
{
        name=commit,
        description={An individual change to a file (or set of files)}
}

\newacronym{ea}{EA}{Electronic Arts}
\newacronym{lm}{LM}{Language Model}
\newacronym{llm}{LLM}{Large Language Model}
\newacronym{ml}{ML}{Machine Learning}
\newacronym{mcqa}{MCQA}{Multiple Choice Question Answering}
\newacronym{nlp}{NLP}{Natural Language Processing}
\newacronym{nl}{NL}{Natural Language}
\newacronym{pl}{PL}{Programming Language}

\def\BibTeX{{\rm B\kern-.05em{\sc i\kern-.025em b}\kern-.08em
    T\kern-.1667em\lower.7ex\hbox{E}\kern-.125emX}}
\begin{document}

\IEEEoverridecommandlockouts
\IEEEpubid{\makebox[\columnwidth]{ 979-8-3503-5067-8/24/\$31.00~\copyright2024 IEEE \hfill}
\hspace{\columnsep}\makebox[\columnwidth]{ }}




\title{Leveraging Large Language Models for Efficient Failure Analysis in Game Development}

\author{
Leonardo Marini$^{1}$,
Linus Gisslén$^{2}$, and
Alessandro Sestini$^{2}$
\\
$^{1}$\textit{Frostbite}, $^{2}$\textit{SEED - \acrfull{ea}}
\\
leonardo.marini@frostbite.com, \{lgisslen, asestini\}@ea.com
}

\maketitle

\IEEEpubidadjcol

\begin{abstract}

In games, and more generally in the field of software development, early detection of bugs is vital to maintain a high quality of the final product.
Automated tests are a powerful tool that can catch a problem earlier in development by executing periodically.
As an example, when new code is submitted to the code base, a new automated test verifies these changes.
However, identifying the specific change responsible for a test failure becomes harder when dealing with batches of changes -- especially in the case of a large-scale project such as a AAA game, where thousands of people contribute to a single code base.
This paper proposes a new approach to automatically identify which change in the code caused a test to fail.
The method leverages Large Language Models (LLMs) to associate error messages with the corresponding code changes causing the failure.
We investigate the effectiveness of our approach with quantitative and qualitative evaluations.
Our approach reaches an accuracy of 71\% in our newly created dataset, which comprises issues reported by developers at \acrshort{ea} over a period of one year.
We further evaluated our model through a user study to assess the utility and usability of the tool from a developer perspective, resulting in a significant reduction in time -- up to 60\% -- spent investigating issues.

\end{abstract}


\section{Introduction}
\label{sec:intro}
In software development, it is crucial to identify and resolve a bug as quickly as possible.
Several factors can impede the early detection of bugs: the size of the code base, the number of contributors involved in writing code and the frequency of their contributions.
Especially in video game development: AAA games are constantly growing in size, resulting in very complex software with a vast amount of interconnected sub-systems, which must communicate and work with each other.
In this scenario, full testing coverage is not achievable by sole manual human intervention.
Automated testing can free up human resources to allow for more meaningful testing at a high level: measuring game balance, difficulty, and potential retention rate.

Automated tests can verify that existing functionalities do not break by running periodically and catching a problem earlier rather than later.
In game development, there are many different automated tests that game development teams may wish to perform.
For example, a usual test in 3D games is to apply a force to an object and verify that it reacts as expected based on its physical properties. Modern AAA games contain numerous objects and multiple physical interactions. Thus, the number of tests can increase exponentially with the size of the game in this context.
Additionally, a game engine like Frostbite, which supports several platforms (e.g. old and new generations of PlayStation and Xbox, Nintendo Switch, and mobile systems like Android and iOS), needs repeating tests to cover the most common hardware and operating systems. These requirements contribute to an increasing number of tests.

The resources and costs needed to execute these tests constrain the frequency of their execution.
Therefore, when new code submissions are too frequent, and the resources become scarce to run tests for every new submission, changes from multiple developers are batched and tested together.
A new problem then arises: how can one tell which one of the changes is the one that contains a bug when a test starts failing?

Nowadays, every game developer should monitor their changes as they are tested and submit a fix if they notice a failure. This task becomes increasingly difficult when dealing with hundreds of tests, without even considering the amount of time required for all of them to complete their execution.
A way to mitigate this issue is to develop systems that  automatically give a notification when a test starts failing.
A naive approach could be to notify every developer who submitted a change when the test started failing.
However, this solution can be noisy when the number of tests and batched \glspl{commit} is large.

\gls{nlp} can be used to understand the meaning of a text \cite{devlin2019bert}. Recently automatic triaging of code bugs with \gls{nlp} is gaining interest in research and industry community: some notable previous works have been conducted by Ubisoft to prevent the introduction of bugs in game development \cite{ubisoft} and by Mozilla to automatically triage bugs, categorizing them based on the title, description, and additional fields filed by the reporting user \cite{CL19}.
Moreover, recent works have shown that these \gls{ml} models can also apply to \glspl{pl} as well, e.g. to understand code \cite{codebert} and translate it into \gls{nl} or other \glspl{pl} \cite{ahmad2021unified}.
Such approaches suggest that it could be possible to analyze errors and \gls{commit} messages, like a human would do. For instance, identifying which one of the \glspl{commit} caused the test to fail. Thus producing a more educated guess and notifying a smaller number of developers who, consequentially, will be able to act faster and fix the bug earlier in development.

In this paper, we propose a method based on BERT \cite{devlin2019bert} that, given an error message as context and multiple descriptions of code changes, can infer the most likely cause of the error. 
By employing this method, we can pinpoint and associate the error message with the specific description of the code change responsible for the test failure, thereby directly notifying the developer who should address the issue. This method effectively reduces the turnaround time for this otherwise cumbersome operation.
Our model achieves an accuracy of 71\% on our newly created dataset, consisting of issues reported by developers of the Frostbite engine that we collected over a year.
We implement the model in a preexisting framework which assists developers in tracking their code submissions.
We assess the quality of the tool with a user study, which revealed that the new approach saves them roughly $41\%\sim60\%$ of the time when investigating the cause of an issue.

To summarize, the contributions of this paper are:

 \begin{itemize}
     \item we introduce a model designed to learn from an error description and multiple commits to identify the one responsible for the error;
     \item we demonstrate the integration of our model into an existing development framework, providing valuable support for professional developers in their daily workflow;
     \item we perform a quantitative analysis comparing various NLP models, as well as a qualitative analysis to evaluate the utility and usability of our integrated approach within the preexisting framework.
 \end{itemize}

The rest of the paper is organized as follows: 
Section \ref{sec:related-work} reviews work from the literature most related to our contributions.
Section \ref{sec:method} formalizes the problem, describing the challenges and the method we used to solve them.
Section \ref{sec:experiments} gives an idea of what data we used to train the model, describes the metrics used for evaluation to compare multiple models, and presents the quantitative results.
We also define the user survey and interpret its qualitative results.
Finally, Section \ref{sec:conclusions} summarizes all the results, list the limitations of the model, and propose a direction for future works.

\section{Related Work}
\label{sec:related-work}
The potential of \gls{nlp} for software development has been gaining interest from both the research and industry communities. Here, we review work from the literature most related to our contributions.

Several approaches employ \gls{nlp} techniques for bug detection and identification. 
\citet{stack_study} empirically demonstrated that developers can benefit significantly from stack trace information during debugging. \citet{security_bug} present a method that utilizes text mining on natural-language descriptions of bug reports to train a statistical model for identifying high-priority security bugs.
CrashDroid \cite{crashdroid} generates reproducible steps by translating the call stack, which contains all method calls from software launch to software crash.
CrashTranslator \cite{crash_translator} advances this concept by automatically reproducing mobile application crashes directly from the stack trace using large language models.
Consequently, numerous research papers have proposed automated approaches to identify the links between crashes and their cause, thereby aiding developers \cite{locating, fault_reside, crashlocator, ubisoft}.

Recently, many \glspl{llm} \cite{devlin2019bert, gpt} have been developed and used specifically for problems related to code and code bases.
Models such as AlphaCode, CodeGen, OctoPack and StarCoder \cite{alphacode, codegen, octopath, starcoder} are examples of \glspl{llm} that have been fine-tuned for program synthesis across multiple programming languages.
At the same time, many encoder-only transformer architectures \cite{transformer_vasani} have been used for code understanding, mainly focusing on code retrieval, classification, and program repair.
Some notable examples are: \citet{cubert} propose a model to obtain a high-quality contextual embedding of source code, mainly used for program understanding; \citet{codebert} propose a bimodal pre-trained model for \glspl{pl}, namely CodeBERT; \citet{pymt5} describe a single model that can both predict whole methods from natural language documentation strings and summarize code into docstrings of any standardized style, specifically for Python; and CodeT5 \cite{codet5}, a unified pre-trained encoder-decoder transformer model that leverages the code semantics conveyed from the developer-assigned identifiers.

To the best of our knowledge, no existing research addresses the problem we have outlined: \emph{given an error message as context and multiple descriptions of code changes, determine the most probable cause of the error among the changes.}
Motivated by the significance of this issue and inspired by the recent success of \glspl{llm} in code generation and interpretation, we present a novel approach for identifying the code change responsible for the error in a specific failed test.
The most relevant related work is CLEVER by \citet{ubisoft}, which employs a two-phase process to intercept risky changes before they reach the central repository.
However, this approach is relatively complex, involving multiple systems communicating with one another.
Furthermore, it does not fulfill our requirements: CLEVER aims to anticipate and intercept a change before it generates an error, whereas our goal is to identify the change that caused the error after the failure has occurred. 

\section{Method}
\label{sec:method}
In this section, we start by defining the problem and the technical terminology used throughout the section.
Then, we present our method to solve the mentioned problem and how developers can interact with it to improve their workflow in triaging errors.

\subsection{Definitions and Preliminaries} 
\label{sec:method_problem}
Given an error message $e$ and $ N_{c}(e) > 1 $ code submissions $c_i$, or \textit{commit}, related to $e$, we want to determine which submission is the most likely cause of the error.
The error message $e$, which consists of a text describing why the test failed, is output by a test designed in such a way as to either succeed or fail.
In the latter, it will output an error message.
As previously mentioned, a usual test in 3D games could verify if an object reacts as expected based on its physical properties when a force is applied to it, while an example of an error message for that test could be: \textit{Testcase: ``TestApplyTerrainDestructionInDSub" asserted with message: Testcase had error in runtime!
}

The test runs continuously: it will gather the latest changes submitted since the test last ran and bundle them together so that $N_{c}(e)$ multiple plausible causes of the failing test are available.
A description $c_i$ accompanies each code submission -- i.e. \gls{commit} message of submission $i$ -- that describes what the change is about, with $i \in [1, N_{c}(e)]$.
This description helps both during code review to understand  what has changed in the code and after to navigate the history of the code base.
Additionally, developers who may want to investigate why a test is failing will typically start by reading the error message and the description of the submitted changes when the failure started occurring before looking into the code itself.

It is worth mentioning that once a test fails, developers can submit new commits, and the test will run again. However, the same test will continue to fail because of the previous commits, independently of the subsequent ones. To test later commits, developers should either (1.) submit another change that fixes the error or (2.) revert the \gls{commit} that caused the first failure.
Later \glspl{commit} could introduce new errors, making it even harder to spot when the latter error was introduced, as it will not manifest until the former is fixed.
As of today, a test that starts failing will send a notification to every developer whose changes were being tested.

We hypothesize that by using \gls{nlp}, we can understand the error message and the description of the changes, thus inferring and selecting the most likely cause of the issue, similarly to what a human developer would do analytically. We formalize this task as a \gls{mcqa} task, where we want to train a model $M_{\theta}(e, c_i)$ that, given an error message $e$ and its related commits $c_i$ outputs a score $s_i$ describing how likely the commit $c_i$ caused the error $e$, with $\theta$ the weights of the model.

\subsection{Model Architecture}
\label{sec:method_our-model}

We decided to use a \acrfull{llm} for our task. An \gls{llm} is a type of \acrfull{lm} designed to predict the likelihood of a sequence of words or tokens in a text.
Unlike traditional \glspl{lm}, \glspl{llm} can pre-process input data, utilizing a vast vocabulary of tokens and handling a significantly higher number of parameters. Thus making them more complex and capable~\cite{gpt-2}.
These models have been pre-trained on extensive text corpora, enabling fine-tuning with a smaller amount of data specific to a given use case.
In our preliminary experiments, we tested multiple LLMs.
We chose BERT \cite{devlin2019bert} as our final model as it outperformed all the others.
We describe the experiments that led us to this choice in Section \ref{sec:experiments}.
This model uses the Transformer architecture \cite{transformer_vasani}, specifically the encoder part with a context window of 512 tokens.
An extra layer was added at the top of the base model to address the specific \gls{mcqa} task.

\subsection{Training the Model}
\label{sec:method_training}

To train the model, as previously mentioned, we formalize the task as an \gls{mcqa} task. We fine-tune a pre-trained BERT model plus our additional layer with a dataset composed of pairs $(e, c)$, where $e$ is an error message and $c$ is the commit that caused it. For each error message, we present the model with four possible \gls{commit} messages $\{c_i \;\; | \;\; i = 1, ..., 4\}$: one that corresponds to the change that caused the error, say $c_k$, plus three more chosen at random from the original dataset. The model $M_\theta(e, c_i)$ then computes four scores $\{s_i \;\; | \;\; i = 1, ..., 4\}$, one for each $c_i$. The optimal model $M_\theta^*$ should output a high score for $c_k$ (the commit that caused the error):
\begin{equation}
\label{eq:opt_model}
    M_\theta^*(e, c_k) \geq M_\theta^*(e, c_i), \;\; i = 1, ..., 4.
\end{equation}
We then use the SoftMax function over the $4$ scores $s_i$ to get the probability $p_i$ that the commit $c_i$ is the one responsible for the error message $e$.
To train the model, we then use the Cross Entropy loss:
\begin{equation}
\label{eq:loss}
    L = - \frac{1}{N} \sum_{b=1}^N \sum_{i = 1}^4 t_i \log(p_i),
\end{equation}
where $t_i$ is $1$ if $i=k$, and $0$ otherwise, and $N$ is the number of data samples in our batch.

Because the model computes a score for each potential choice, it remains independent of the number of commits $N_{c}(e)$ related to an error message during inference. This factor is of fundamental importance during deployment as it means that the model can accept more -- or eventually less -- than four commits as input and still be able to guess which commit caused the error. The highest score among the possible options determines the final choice of the model. The decision to model the problem as an \gls{mcqa} task stems from the fact that in real-world scenarios, there are often multiple submissions per failed test, and usually only one of them is the root cause of the reported error message in the output.

\subsection{Implementation of the tool}
\label{sec:method_implementation}

\begin{figure*}
    \begin{center}
        \includegraphics[width=0.85\textwidth]{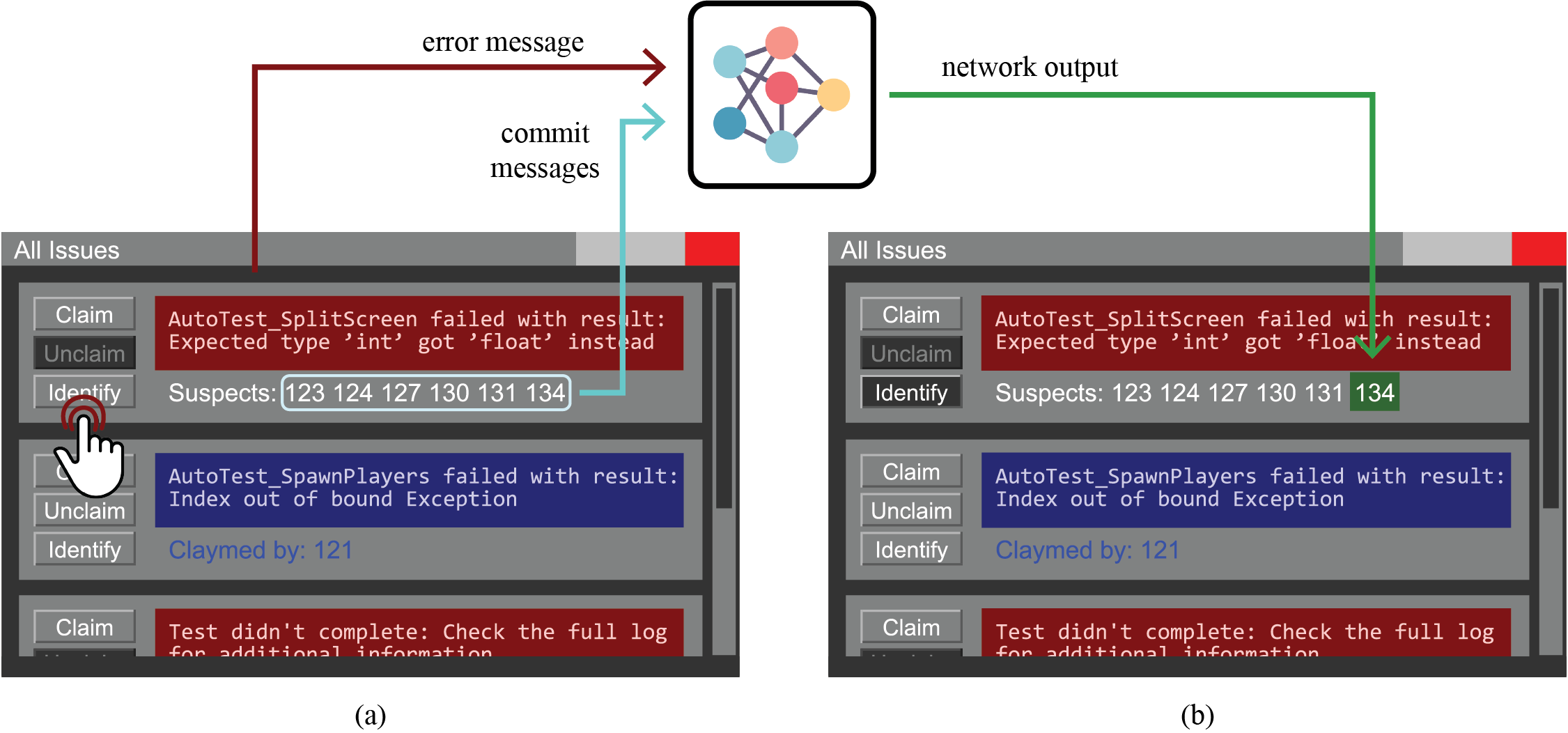}
    \end{center}
    \caption{Mock-up of the tool showing the main existing features and the additional ``\textit{Identify}'' button (in (a)). Each entry in the view represents a failed test. At the top of each entry, we can see the error message $e$. Below the message, the framework shows a list of IDs: each ID corresponds to a commit $c_i$. Once a user presses the identify button, the model will execute, and the estimated ID that caused the error will be highlighted (i.e., number 134 in (b)). Once a user identifies the correct ID, they can claim it, and the view will display \textit{``Claimed by: ID''}. }
    \label{fig:mock}
\end{figure*}

To evaluate the method, we made it available to developers by incorporating it into a preexisting framework. This new feature allowed for an easier transition without changing too much of the workflow that the developers are used to. Figure \ref{fig:mock} shows a simplified visualization of the framework.

The framework keeps track of every change submitted to the code base, which means that for each submission, the framework updates a record corresponding to the commit with all the tests that pass and fail.
This way, a developer can monitor the progress of their commit as it completes the tests, which are initially all in a pending state and will be updated to either pass or fail.
When all the tests have finished running, the developer knows how well their change performed based on this metric. 
Additionally, developers will receive a notification as soon as one of their changes causes a new breakage (i.e. a test that started failing when testing a batch of changes that included their change), so they do not need to wait for all the tests to finish running to take action and start working on a fix.

All of the errors are collected and presented in a secondary view (Figure~\ref{fig:mock}(a) shows an example) that provides every user access to all of the issues occurring on the test farm.
Each issue includes a list of suspects, namely the changes that were submitted when the test started failing.
Usually, a user would attempt to resolve the issue by reading the error message and then going through all the suspects, including the description of all the commits until they find one that is most likely to have caused the error. As previously mentioned, this process is time-consuming and prone to possible errors since a user needs to read all the commits for all the issues.

To speed up the process and reduce errors, we added the feature as a button to every issue appearing in the secondary interface that will send a request to identify the most likely cause of that issue when clicked (Figure~\ref{fig:mock}(a)).
Once the model completes the prediction, it will highlight the corresponding change (Figure~\ref{fig:mock}(b)).
The request sent to the model contains the error message $e$ relative to that particular issue and all the \gls{commit} messages $c_i$ associated with that error.
As mentioned, our training methodology makes the model entirely independent of the number of $c_i$ inputs because the model computes a score for each commit rather than a probability.
This design allows the model to accept more than the standard four commits used during training.
The higher the score, the more likely that particular commit caused that issue. With this information, the framework automatically highlights the primary suspect to developers. 
This way, the user gets a suggestion of the most likely suspect that may not correspond to the truth, and he is free to ignore the suggestion and analyze the rest of the suspects.

\section{Experiments}
\label{sec:experiments}
This section presents the data used to conduct the experiments and the methodology employed to collect such data.
We then detail the metrics used to measure the quantitative results, followed by a presentation of the results comparing the different LLMs employed in this study.
Finally, we showcase the results gathered from a user study we conducted to assess the value of our method as a tool to assist developers.

\subsection{Data}
\label{sec:experiments_data}
We collect a dataset of historic errors via the framework mentioned in Section~\ref{sec:method_implementation}.
For every new issue appearing on the tool, developers can claim any such issue if they think they caused it, or specify a \gls{commit} if they are sure about which one caused it.
We select only those records that include the faulty commit.
All collected records were previously reported manually by developers.
Thus, we can infer that the labels that we use for training are correct.
We collected $n \approx 2500$ records matching these requirements over the past year.

Each record $i$, with $i=1, ..., n$, in the dataset is composed of an error message $e_i$, the description of the \gls{commit} (i.e. \gls{commit} message) $c_i$ that caused it, three additional commit messages ($c_j \; | \; j \in [1,n] \wedge j \neq i$) picked at random from the same dataset -- to compose an \gls{mcqa} task -- and a label $y \in [1,4]$ to identify which one of the \glspl{commit} is the culprit of the error.
Each pair $\{(e_i, c_k) \;\; | \;\; k=[1,4]\}$ was truncated to 512 tokens before being input into the model, as that is a limitation imposed by the base BERT model.
It is important to note that we construct the dataset having the commit that caused the issue at a random position within the range of $[1, 4]$.
Specifically, in the final dataset, the correct answer appears an equal number of times across all positions.

Unfortunately, the data we collected is proprietary and may not be shared. However, we provide some representative examples in Appendix~\ref{sec:appendix} to enable others to reproduce our results.

\subsection{Experimental Setup}
In all the experiments, each model was trained for three epochs, with a learning rate of \num{5e-5} and batch size of 8. We split the dataset composed of $n \approx 2500$ records into three subsets: training, validation and test with a ratio of 80\%:10\%:10\%. For each model, we perform early stopping based on the accuracy of the validation set.

\subsection{Quantitative Evaluation}
For our quantitative analyses, we use an accuracy metric to evaluate the performance of the models on our dataset. More precisely, the proportion of correctly guessed examples over the total number of examples. As mentioned in Section \ref{sec:experiments_data}, each sample includes four possible candidates which may have caused the error. Thus, we can expect a model that gives random guesses to attain a score of 25\%. Another valuable metric is the time saved for finding the correct cause. We will analyze this metric later in the document, in Section \ref{sec:experiments_user-study}.


\begin{table}
    \caption{Model Comparison. This table highlights the performances of different tested models in terms of training loss, evaluation loss -- the loss during the validation phase -- and general accuracy using the test set. As the table shows, the BERT model outperforms all the others, making it the best-performing language model among those tested. A down arrow indicates that lower values are better, while an up arrow indicates that higher values are preferred. The \textit{Random Agent} baseline in the last row represents the performance achieved by a random guess.}
    \begin{center}
    \begin{tabular}{l|r|r|r}
                & Training Loss $\downarrow$ & Evaluation Loss $\downarrow$ & Accuracy $\uparrow$ \\
        \hline
        \textbf{BERT}~\cite{devlin2019bert} & \textbf{0.72} &\textbf{ 1.11} & \textbf{0.71} \\
        XLNet~\cite{xlnet}                  & 1.41          & 1.39          & 0.31 \\
        BigBird~\cite{bigbird}              & 1.39          & 1.38          & 0.31 \\
        MEGA~\cite{mega}                    & 1.39          & 1.39          & 0.22 \\
        Random Agent                        & --            & --            & 0.25 \\
        \hline
    \end{tabular}
    \label{tab:model-comparison}
    \end{center}
\end{table}

Table \ref{tab:model-comparison} shows a comparison of the pre-trained models we experimented with -- BERT~\cite{devlin2019bert}, XLNet~\cite{xlnet}, BigBird~\cite{bigbird}, and MEGA~\cite{mega} -- in terms of training loss, evaluation loss, and accuracy.
The first value of each row is the final training loss at the end of the $3^{rd}$ and last epoch.
The latter two numbers of each row represent the loss on the validation set and the accuracy in the test set based on the epoch in which the model performed best with the validation one.
The loss function, shown in Equation \ref{eq:loss}, is the inverse of the score and what we want to minimize.
In contrast, the accuracy is the metric used to rank the models and what we want to maximize.
Note how all the models perform similarly, except for BERT, which vastly outperforms the others: it is more than twice as accurate (accuracy $= 71\%$) compared to the others (accuracy $\leq 31\%$).
All hyperparameters were chosen after a set of preliminary experiments made with different configurations for each model.
Of note is that all the tested models except for BERT outperformed the random agent baseline only slightly, and MEGA performed even less effectively.
During training, these models demonstrated a rapid overfitting to the training dataset, which decreased their generalization performance to the test set.

\renewcommand{\arraystretch}{2}
\begin{table*}
\centering
\caption{Summary of participants for the user study. All participants are experts in the tool, and one of the main duties in their daily workflow is to manually identify the cause of an error in case of a test failure. The number reported in this table represents the mean and standard deviation of the answer received during the user study.}
\begin{tabular}{l|cccc}
\hline
& \textbf{\makecell{Time inspecting \\ issue}} & \textbf{Experience} & \textbf{\makecell{Machine learning \\ knowledge}} & \textbf{\makecell{Frequency of use \\ of the tool}} \\ \hline
Average & 16.0 $\pm$ 4.18 minutes & 2.6 $\pm$ 3.57 years &  $\sim$ Medium & $\sim$ Multiple times per day \\
\hline
\end{tabular}
\label{tab:participants}
\end{table*}

\renewcommand{\arraystretch}{1.5}
\begin{table*}
\centering
\caption{Results of some of the most important answers of the user study. Most questions expected a value between 1 and 5 as an answer. For question number 2, the participants could answer either with an interval of 20\% between 1\% and 100\% (e.g. 41\% - 60\%) or 0\%. $\mu$ represents the mean of the values answered by participants; $\sigma$ is the standard deviation of the answers; and Mo is the mode of these values.}
\begin{tabular}{r|l|c|c|c}
\hline
\textbf{Nr.} & \textbf{Question} & $\mu$ & $\sigma$ & Mo \\ \hline
1 & How useful did you find this new feature? & 4.6 & $\pm$0.55 & 5 \\ 
2 & Based on how much time you used to spend on investigating issues, how much time did you save? & $41\% \sim $60\% & - & - \\
3 & Did you feel the need to double check the identified suspect before assigning an issue? & 3.8 & $\pm$0.84 & 4 \\
4 & Do you find it valuable to have the framework suggest you one most likely suspect, even when it is wrong? & 4.4 & $\pm$0.89 & 5 \\
5 & How often did you end up using the model compared to what was already available to you? & 4.2 & $\pm$0.84 & 4 \\
6 & Would you like to keep using this new feature? & 5.0 & $\pm$0.00 & 5 \\
\hline
\end{tabular}
\label{tab:study_results}
\end{table*}

\renewcommand{\arraystretch}{0.5}
\begin{figure}
    \begin{center}
        \begin{tabular}{c}
        \includegraphics[width=0.85\columnwidth]{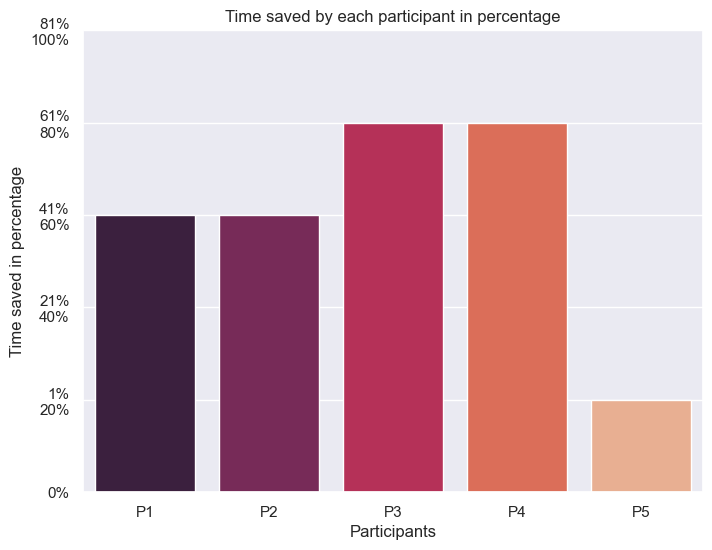} \\ (a) \\ \\
        \includegraphics[width=0.85\columnwidth]{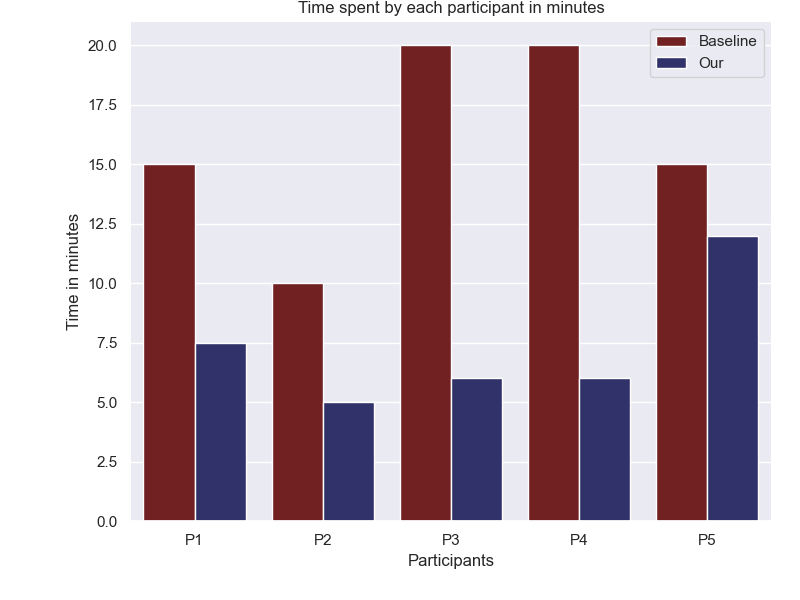}\\ (b) \\
    \end{tabular}
    \end{center}
    \caption{Time saved by each participant. In (a), the percentage indicates the amount of time saved by each participant compared to the previous manual approach. In (b), the time spent investigating an issue with the previous manual approach and after our framework. For both plots, the X-axis represents the participants (e.g. P1 means participant 1).}
    \label{fig:time_saved}
\end{figure}

\subsection{User Study}
\label{sec:experiments_user-study}
In order to assess the usability and effectiveness of the proposed method, we carried out a user study utilizing a survey that consisted of 17 closed-ended questions accompanied by an optional open-ended question for additional feedback. The majority of the questions employed a Likert scale ranging from 1 to 5 for responses, while the remaining questions featured multiple-choice answer options. The survey was divided into four distinct sections, as follows:


\begin{itemize}
    \item 3 questions about the user's background
    \item 5 questions about the user's current workflows
    \item 7 questions about our new proposed method
    \item 3 questions about future improvements
\end{itemize}

The purpose of the first part is to put the rest of the answer in perspective of the user's prior knowledge:  senior developers could be less keen on changing their workflow simply due to the fact that they have gotten used to working in a certain way for a long time. These questions ask about the user's role and seniority level within the company, and level of knowledge of \gls{ml}.

The following five questions probe how familiar the user is with using the existing framework.
Developers who have never used a particular feature will likely give a neutral answer to later questions that compare the new workflow to the existing one. 
These questions help us identify those answers whose score is irrelevant because the user would have skipped if they could.
Thus, we can exclude from the final result to prevent altering the average score over the entire population.
The third part of the survey evaluates the quality of the proposed method.
The questions range from requesting a personal evaluation of the user's experience to more detailed questions that assess the magnitude of the benefits of the proposed approach, such as time saved.
Finally, we ask the user for potential feedback to gather and prioritize ideas later mentioned in Section \ref{sec:conclusions} for future works.
Five in-house professional developers volunteered to try the new tool and completed our user study.
Although they represent a small sample, they were the perfect candidates for this study as they are the primary users of the tool and the principal developers who look at the error messages and manually assign a commit to the error.
As shown in Table~\ref{tab:participants}, the knowledge level of the tool is high, with participants using the tool several times during the day.
On average, each user spends at least 15 minutes detecting the commit that caused one error.
However, their \gls{ml} knowledge ranges from low to medium.
The users tested the proposed framework over a period of 30 days.
In our qualitative evaluation, we assume that each user is 100\% accurate in identifying the correct commit.
Any amount of time saved implies that the tool was successful in pointing them to the correct commit in less time than what developers typically require.

Table~\ref{tab:study_results} illustrates the main results of this user study.
Overall, the proposed method was very well received, with significant improvements and minimal drawbacks.
With our approach, the turnaround time dropped from an average of 16 minutes -- as shown in Table~\ref{tab:participants} -- to approximately 8 minutes: an improvement of $41\%\sim60\%$ as reported in Table~\ref{tab:study_results}.
In Figure \ref{fig:time_saved} we illustrate the time saved for each of the participants.
All participants found the approach very useful, and all of them would be willing to continue using it in the future.
This is even though the participants felt the need to verify the identified suspect after our model suggestion.
It is interesting to note, as the table shows, that the suggestions were still useful, and the participants found the tool valuable, even when the model was wrong.
We would also like to emphasize that developers repeat this process multiple times during the day.
Consequently, although the time saved on a singular issue may not appear substantial, the cumulative effect over a day's work can result in a considerable reduction in time.
Of particular note are the answers from Participant 5 (P5 in Figure~\ref{fig:time_saved}). This developer is the one who saved less time (between 1\%$\sim$20\%), but he expresses a high desire to continue using the approach (5/5 for question 6 in Table~\ref{tab:study_results}).
We motivate these answers with a quote from the same participant: ``\textit{It was quite nice that the framework highlighted the most probable cause of breakage. However, I still had to double-check if it was correct.
Also because even though I love automation and machine learning I have some trust issues and I rather double-check carefully}''.

\section{Conclusions and Limitations}
\label{sec:conclusions}

In this paper, we propose a practical method to improve software development efficiency by assisting developers in tracking changes that cause issues in the code base manifesting as failing tests.
Our model achieves an accuracy of 71\% over the dataset of real issues we collected.
A simple user study shows that the proposed method resulted in a significant reduction in time spent investigating failures: up to 60\% of time saved to investigate a failed test and identify its cause. Moreover, the approach does not bring any relevant drawbacks to the existing framework.

While our tests demonstrate the good performance of our approach, the framework suffers from some limitations that will be addressed in future works. The input to our model is limited: 512 tokens are easily filled with the error message and one description of a change.
With the possibility of using  larger inputs, it would be interesting to see if the model performs better by including more information, such as the name of the failed test, a description of what it tests, the names of the files that were changed, as well as the differences between files.
Another interesting approach to consider involves the abbreviation of error messages. Given that these messages frequently consist of similar syntax, a potential area for future exploration could involve the automatic truncation of these messages without sacrificing meaning.

There are many things that future research could improve upon our work. As mentioned in Section \ref{sec:method_training}, we wanted our model to resemble the real-case scenario as closely as possible.
However, there are limitations to doing so.
For instance, we assume that for every error occurring in a test, there is always one, and one only, \gls{commit} that introduced it.
In reality, some of the errors could have been generated by a fault in the machine that was running it, or even that the test was not designed correctly and its result is non-deterministic.
In these cases, there would not be any submission causing the error, and our model does not account for that.

\bibliographystyle{IEEEtranN}
{\footnotesize \bibliography{main}}

\begin{thebibliography}{25}
\providecommand{\natexlab}[1]{#1}
\providecommand{\url}[1]{#1}
\csname url@samestyle\endcsname
\providecommand{\newblock}{\relax}
\providecommand{\bibinfo}[2]{#2}
\providecommand{\BIBentrySTDinterwordspacing}{\spaceskip=0pt\relax}
\providecommand{\BIBentryALTinterwordstretchfactor}{4}
\providecommand{\BIBentryALTinterwordspacing}{\spaceskip=\fontdimen2\font plus
\BIBentryALTinterwordstretchfactor\fontdimen3\font minus \fontdimen4\font\relax}
\providecommand{\BIBforeignlanguage}[2]{{%
\expandafter\ifx\csname l@#1\endcsname\relax
\typeout{** WARNING: IEEEtranN.bst: No hyphenation pattern has been}%
\typeout{** loaded for the language `#1'. Using the pattern for}%
\typeout{** the default language instead.}%
\else
\language=\csname l@#1\endcsname
\fi
#2}}
\providecommand{\BIBdecl}{\relax}
\BIBdecl

\bibitem[Devlin et~al.(2019)Devlin, Chang, Lee, and Toutanova]{devlin2019bert}
J.~Devlin, M.-W. Chang, K.~Lee, and K.~Toutanova, ``Bert: Pre-training of deep bidirectional transformers for language understanding,'' 2019.

\bibitem[Nayrolles and Hamou-Lhadj(2018)]{ubisoft}
M.~Nayrolles and A.~Hamou-Lhadj, ``Clever: Combining code metrics with clone detection for just-in-time fault prevention and resolution in large industrial projects,'' in \emph{2018 IEEE/ACM 15th International Conference on Mining Software Repositories (MSR)}.\hskip 1em plus 0.5em minus 0.4em\relax Gothenburg, Sweden: IEEE, 2018, pp. 153--164.

\bibitem[Castelluccio and Ledru(2019)]{CL19}
\BIBentryALTinterwordspacing
M.~Castelluccio and S.~Ledru. (2019) Teaching machines to triage firefox bugs. Mozilla. [Online]. Available: \url{https://hacks.mozilla.org/2019/04/teaching-machines-to-triage-firefox-bugs/}
\BIBentrySTDinterwordspacing

\bibitem[Feng et~al.(2020)Feng, Guo, Tang, Duan, Feng, Gong, Shou, Qin, Liu, Jiang, et~al.]{codebert}
Z.~Feng, D.~Guo, D.~Tang, N.~Duan, X.~Feng, M.~Gong, L.~Shou, B.~Qin, T.~Liu, D.~Jiang \emph{et~al.}, ``Codebert: A pre-trained model for programming and natural languages,'' \emph{arXiv preprint arXiv:2002.08155}, 2020.

\bibitem[Ahmad et~al.(2021)Ahmad, Chakraborty, Ray, and Chang]{ahmad2021unified}
W.~U. Ahmad, S.~Chakraborty, B.~Ray, and K.-W. Chang, ``Unified pre-training for program understanding and generation,'' 2021.

\bibitem[Schroter et~al.(2010)Schroter, Schr{\"o}ter, Bettenburg, and Premraj]{stack_study}
A.~Schroter, A.~Schr{\"o}ter, N.~Bettenburg, and R.~Premraj, ``Do stack traces help developers fix bugs?'' in \emph{2010 7th IEEE working conference on mining software repositories (MSR 2010)}.\hskip 1em plus 0.5em minus 0.4em\relax IEEE, 2010, pp. 118--121.

\bibitem[Gegick et~al.(2010)Gegick, Rotella, and Xie]{security_bug}
M.~Gegick, P.~Rotella, and T.~Xie, ``Identifying security bug reports via text mining: An industrial case study,'' in \emph{2010 7th IEEE Working Conference on Mining Software Repositories (MSR 2010)}.\hskip 1em plus 0.5em minus 0.4em\relax IEEE, 2010, pp. 11--20.

\bibitem[White et~al.(2015)White, Linares-V{\'a}squez, Johnson, Bernal-C{\'a}rdenas, and Poshyvanyk]{crashdroid}
M.~White, M.~Linares-V{\'a}squez, P.~Johnson, C.~Bernal-C{\'a}rdenas, and D.~Poshyvanyk, ``Generating reproducible and replayable bug reports from android application crashes,'' in \emph{2015 IEEE 23rd International Conference on Program Comprehension}.\hskip 1em plus 0.5em minus 0.4em\relax IEEE, 2015, pp. 48--59.

\bibitem[Huang et~al.(2023)Huang, Wang, Liu, Wang, Wang, Chen, Hu, and Wang]{crash_translator}
Y.~Huang, J.~Wang, Z.~Liu, Y.~Wang, S.~Wang, C.~Chen, Y.~Hu, and Q.~Wang, ``Crashtranslator: Automatically reproducing mobile application crashes directly from stack trace,'' \emph{arXiv preprint arXiv:2310.07128}, 2023.

\bibitem[Gong et~al.(2014)Gong, Zhang, Seo, and Kim]{locating}
L.~Gong, H.~Zhang, H.~Seo, and S.~Kim, ``Locating crashing faults based on crash stack traces,'' \emph{arXiv preprint arXiv:1404.4100}, 2014.

\bibitem[Gu et~al.(2019)Gu, Xuan, Zhang, Zhang, Fan, Xie, and Qian]{fault_reside}
Y.~Gu, J.~Xuan, H.~Zhang, L.~Zhang, Q.~Fan, X.~Xie, and T.~Qian, ``Does the fault reside in a stack trace? assisting crash localization by predicting crashing fault residence,'' \emph{Journal of Systems and Software}, vol. 148, pp. 88--104, 2019.

\bibitem[Wu et~al.(2014)Wu, Zhang, Cheung, and Kim]{crashlocator}
R.~Wu, H.~Zhang, S.-C. Cheung, and S.~Kim, ``Crashlocator: Locating crashing faults based on crash stacks,'' in \emph{Proceedings of the 2014 International Symposium on Software Testing and Analysis}, 2014, pp. 204--214.

\bibitem[Achiam et~al.(2023)Achiam, Adler, and et~al.]{gpt}
\BIBentryALTinterwordspacing
O.~J. Achiam, S.~Adler, and S.~A. et~al., ``Gpt-4 technical report,'' 2023. [Online]. Available: \url{https://api.semanticscholar.org/CorpusID:257532815}
\BIBentrySTDinterwordspacing

\bibitem[Li et~al.(2022)Li, Choi, Chung, Kushman, Schrittwieser, Leblond, Eccles, Keeling, Gimeno, Dal~Lago, et~al.]{alphacode}
Y.~Li, D.~Choi, J.~Chung, N.~Kushman, J.~Schrittwieser, R.~Leblond, T.~Eccles, J.~Keeling, F.~Gimeno, A.~Dal~Lago \emph{et~al.}, ``Competition-level code generation with alphacode,'' \emph{Science}, vol. 378, no. 6624, pp. 1092--1097, 2022.

\bibitem[Nijkamp et~al.(2022)Nijkamp, Pang, Hayashi, Tu, Wang, Zhou, Savarese, and Xiong]{codegen}
E.~Nijkamp, B.~Pang, H.~Hayashi, L.~Tu, H.~Wang, Y.~Zhou, S.~Savarese, and C.~Xiong, ``Codegen: An open large language model for code with multi-turn program synthesis,'' \emph{arXiv preprint arXiv:2203.13474}, 2022.

\bibitem[Muennighoff et~al.(2023)Muennighoff, Liu, Zebaze, Zheng, Hui, Zhuo, Singh, Tang, Von~Werra, and Longpre]{octopath}
N.~Muennighoff, Q.~Liu, A.~Zebaze, Q.~Zheng, B.~Hui, T.~Y. Zhuo, S.~Singh, X.~Tang, L.~Von~Werra, and S.~Longpre, ``Octopack: Instruction tuning code large language models,'' \emph{arXiv preprint arXiv:2308.07124}, 2023.

\bibitem[Li et~al.(2023)Li, Allal, Zi, Muennighoff, Kocetkov, Mou, Marone, Akiki, Li, Chim, et~al.]{starcoder}
R.~Li, L.~B. Allal, Y.~Zi, N.~Muennighoff, D.~Kocetkov, C.~Mou, M.~Marone, C.~Akiki, J.~Li, J.~Chim \emph{et~al.}, ``Starcoder: may the source be with you!'' \emph{arXiv preprint arXiv:2305.06161}, 2023.

\bibitem[Vaswani et~al.(2017)Vaswani, Shazeer, Parmar, Uszkoreit, Jones, Gomez, Kaiser, and Polosukhin]{transformer_vasani}
A.~Vaswani, N.~Shazeer, N.~Parmar, J.~Uszkoreit, L.~Jones, A.~N. Gomez, {\L}.~Kaiser, and I.~Polosukhin, ``Attention is all you need,'' \emph{Advances in neural information processing systems}, vol.~30, 2017.

\bibitem[Kanade et~al.(2020)Kanade, Maniatis, Balakrishnan, and Shi]{cubert}
A.~Kanade, P.~Maniatis, G.~Balakrishnan, and K.~Shi, ``Learning and evaluating contextual embedding of source code,'' in \emph{International conference on machine learning}.\hskip 1em plus 0.5em minus 0.4em\relax PMLR, 2020, pp. 5110--5121.

\bibitem[Clement et~al.(2020)Clement, Drain, Timcheck, Svyatkovskiy, and Sundaresan]{pymt5}
C.~B. Clement, D.~Drain, J.~Timcheck, A.~Svyatkovskiy, and N.~Sundaresan, ``Pymt5: multi-mode translation of natural language and python code with transformers,'' \emph{arXiv preprint arXiv:2010.03150}, 2020.

\bibitem[Wang et~al.(2021)Wang, Wang, Joty, and Hoi]{codet5}
Y.~Wang, W.~Wang, S.~Joty, and S.~C. Hoi, ``Codet5: Identifier-aware unified pre-trained encoder-decoder models for code understanding and generation,'' \emph{arXiv preprint arXiv:2109.00859}, 2021.

\bibitem[Radford et~al.(2019)Radford, Wu, Child, Luan, Amodei, Sutskever, et~al.]{gpt-2}
A.~Radford, J.~Wu, R.~Child, D.~Luan, D.~Amodei, I.~Sutskever \emph{et~al.}, ``Language models are unsupervised multitask learners,'' \emph{OpenAI blog}, vol.~1, no.~8, p.~9, 2019.

\bibitem[Yang et~al.(2019)Yang, Dai, Yang, Carbonell, Salakhutdinov, and Le]{xlnet}
Z.~Yang, Z.~Dai, Y.~Yang, J.~Carbonell, R.~R. Salakhutdinov, and Q.~V. Le, ``Xlnet: Generalized autoregressive pretraining for language understanding,'' \emph{Advances in neural information processing systems}, vol.~32, 2019.

\bibitem[Zaheer et~al.(2020)Zaheer, Guruganesh, Dubey, Ainslie, Alberti, Ontanon, Pham, Ravula, Wang, Yang, et~al.]{bigbird}
M.~Zaheer, G.~Guruganesh, K.~A. Dubey, J.~Ainslie, C.~Alberti, S.~Ontanon, P.~Pham, A.~Ravula, Q.~Wang, L.~Yang \emph{et~al.}, ``Big bird: Transformers for longer sequences,'' \emph{Advances in neural information processing systems}, vol.~33, pp. 17\,283--17\,297, 2020.

\bibitem[Ma et~al.(2022)Ma, Zhou, Kong, He, Gui, Neubig, May, and Zettlemoyer]{mega}
X.~Ma, C.~Zhou, X.~Kong, J.~He, L.~Gui, G.~Neubig, J.~May, and L.~Zettlemoyer, ``Mega: moving average equipped gated attention,'' \emph{arXiv preprint arXiv:2209.10655}, 2022.

\end{thebibliography}

\appendix
\label{sec:appendix}
Here we provide two examples of failed tests, plus some potential commits that could be the cause of the failed tests. 

\begin{figure*}
\label{fig:failed_test_1}
\begin{tcolorbox}[width=\textwidth, title={Error Message}, colframe=Maroon]
Testcase: ``AutoTest\_SplitScreen'' asserted with message: Testcase AutoTest\_SplitScreen failed with result: Failed - (Please set FailureMessage on TestCaseEntity to provide reason)    
\end{tcolorbox}
\begin{tcbraster}[ raster columns=2, raster equal height, raster width=\textwidth]
\begin{tcolorbox}[title={Commit 1}]
{[ES]}  Implement support for ShaderBlendMode\_PremultipliedColor Resolves ERROR-192388  Add missing transmittance input to lit root node Resolves
\end{tcolorbox}
\begin{tcolorbox}[title={Commit 2}, colframe=LimeGreen]
{[CharacterPhysics]} Replace terrain in Autotest levels with a large ground box as it is unnecessary (and causing failures on some IOS devices)
\end{tcolorbox}
\end{tcbraster}
\begin{tcbraster}[ raster columns=2, raster equal height, raster width=\textwidth]
\begin{tcolorbox}[title={Commit 3}]
Move MixinRuntimeComponent to an internal RuntimeVariations detail, refactor internals to capture individual variant layers' entity ranges. 
\end{tcolorbox}
\begin{tcolorbox}[title={Commit 4}]
{[Localization]} Timestamp Formatter Entity
\end{tcolorbox}
\end{tcbraster}

\begin{center}
(a)
\end{center}

\begin{tcolorbox}[width=\textwidth, title={Error Message}, colframe=Maroon]
{Assert: (m\_meshDeformerMap.find(serializationType) != m\_meshDeformerMap.cend()) \\ (m\_meshDeformerMap.find(serializationType) != m\_meshDeformerMap.cend())}
\end{tcolorbox}
\begin{tcbraster}[ raster columns=2, raster equal height, raster width=\textwidth]
\begin{tcolorbox}[title={Commit 1}]
[MeshOperate] LodGenerator: Improved shadow mesh generation.  Added a separate container of shadow LODs per submesh. Whereas the existing LOD collection of a component contains conventional LODs for use in building conventional LOD output meshes.
\end{tcolorbox}
\begin{tcolorbox}[title={Commit 2}]
[Movie] Fix dependency issues for Movie source data modules - Updated movie screenshots as background has changed.
\end{tcolorbox}
\end{tcbraster}
\begin{tcbraster}[ raster columns=2, raster equal height, raster width=\textwidth]
\begin{tcolorbox}[title={Commit 3}]
[Appereance] Added Generic Recipe Item and  RecipeItemComposer/ Decomposer Entities Reviewed by A. Reds.
\end{tcolorbox}
\begin{tcolorbox}[title={Commit 4}, colframe=LimeGreen]
[MeshDeformer] Add support for multiple types of GPU compute deformers.  Simply adding new deformers to MeshStream would be a lot less code. However, that would mean that World.Base would have to depend on all types of GPU compute deformers we'd like.
\end{tcolorbox}
\end{tcbraster}
\begin{center}
(b)
\end{center}

\caption{Two examples of error messages and relative commits. In red is the error message, and in green is the commit that caused the error. The model takes as input the couple $(e, c_i)$ for each $c_i$ and outputs a score indicating how probable that commit caused the error message.}
\end{figure*}
\end{document}